\documentclass[10pt,twocolumn,letterpaper]{article}

\usepackage[pagenumbers]{cvpr} 

\usepackage{multirow}
\usepackage{multicol}
\usepackage{graphicx}
\usepackage{amsmath}
\usepackage{amssymb}
\usepackage{booktabs}
\usepackage[pagebackref,breaklinks,colorlinks]{hyperref}
\usepackage{marvosym}
\usepackage{ifsym}

\usepackage[capitalize]{cleveref}
\crefname{section}{Sec.}{Secs.}
\Crefname{section}{Section}{Sections}
\Crefname{table}{Table}{Tables}
\crefname{table}{Tab.}{Tabs.}

\newcommand\blfootnote[1]{%
  \begingroup
  \renewcommand\thefootnote{}\footnote{#1}%
  \addtocounter{footnote}{-1}%
  \endgroup
}

\begin{document}

\title{BEVFormer v2: Adapting Modern Image Backbones to \\ Bird's-Eye-View Recognition via Perspective Supervision}

\author{Chenyu Yang\textsuperscript{\rm 1*} ~~ 
Yuntao Chen\textsuperscript{\rm 2*} ~~ 
Hao Tian\textsuperscript{\rm 3*} ~~ 
Chenxin Tao\textsuperscript{\rm 1} ~~ 
Xizhou Zhu\textsuperscript{\rm 3} ~~ 
Zhaoxiang Zhang\textsuperscript{\rm 2,4} \\
Gao Huang\textsuperscript{\rm 1} ~~ 
Hongyang Li\textsuperscript{\rm 5} ~~
Yu Qiao\textsuperscript{\rm 5} ~~
Lewei Lu\textsuperscript{\rm 3} ~~ 
Jie Zhou\textsuperscript{\rm 1} ~~
Jifeng Dai\textsuperscript{\rm 1,5\Letter}
\\ [0.15cm]
\textsuperscript{\rm 1}Tsinghua University ~~ 
\textsuperscript{\rm 2}Centre for Artificial Intelligence and Robotics, HKISI\_CAS \\ 
\textsuperscript{\rm 3}SenseTime Research ~~
\textsuperscript{\rm 4}Institute of Automation, Chinese Academy of Science (CASIA) \\
\textsuperscript{\rm 5}Shanghai Artificial Intelligence Laboratory
\\ [0.15cm]
{\tt\small \{yangcy19, tcx20\}@mails.tsinghua.edu.cn, chenyuntao08@gmail.com, tianhao2@senseauto.com}\\
{\tt\small \{zhuwalter, luotto\}@sensetime.com, zhaoxiang.zhang@ia.ac.cn}\\
{\tt\small \{gaohuang, jzhou, daijifeng\}@tsinghua.edu.cn, \{lihongyang, qiaoyu\}@pjlab.org.cn}
}

\maketitle

\begin{abstract}
We present a novel bird's-eye-view (BEV) detector with perspective supervision, which converges faster and better suits modern image backbones.
Existing state-of-the-art BEV detectors are often tied to certain depth pre-trained backbones like VoVNet, hindering the synergy between booming image backbones and BEV detectors.
To address this limitation, we prioritize easing the optimization of BEV detectors by introducing perspective view supervision.
To this end, we propose a two-stage BEV detector, where proposals from the perspective head are fed into the bird's-eye-view head for final predictions.
To evaluate the effectiveness of our model, we conduct extensive ablation studies focusing on the form of supervision and the generality of the proposed detector.
The proposed method is verified with a wide spectrum of traditional and modern image backbones and achieves new SoTA results on the large-scale nuScenes dataset.
The code shall be released soon.
\end{abstract}
\blfootnote{*: Equal contribution.}
\blfootnote{\Letter: Corresponding author.}

\vspace{-5pt}
\section{Introduction}
Bird's-eye-view(BEV) recognition models~\cite{OFT,pseudo-lidar,VPN,LSS,bevformer,CVT,PETR} have attracted interest in autonomous driving as they can naturally integrate partial raw observations from multiple sensors into a unified holistic 3D output space.
A typical BEV model is built upon an image backbone, followed by a view transformation module that lifts perspective image features into BEV features, which are further processed by a BEV feature encoder and some task-specific heads.
Although much effort is put into designing the view transformation module~\cite{CVT,LSS,bevformer} and incorporating an ever-growing list of downstream tasks~\cite{LSS,fiery} into the new recognition framework, the study of image backbones in BEV models receives far less attention.
As a cutting-edge and highly demanding field, it is natural to introduce modern image backbones into autonomous driving.
Surprisingly, the research community chooses to stick with VoVNet~\cite{Vovnet} to enjoy its large-scale depth pre-training~\cite{DD3D}.
In this work, we focus on unleashing the full power of modern image feature extractors for BEV recognition to unlock the door for future researchers to explore better image backbone design in this field. 

However, simply employing those modern image backbones without proper pre-training fails to yield satisfactory results. 
For instance, an ImageNet \cite{ImageNet} pre-trained ConvNeXt-XL \cite{ConvNext} backbone performs just on par with a DDAD-15M pre-trained VoVNet-99 \cite{DD3D} for 3D object detection, albeit the latter has 3.5$\times$ parameters of the former. 
We owe the struggle of adapting modern image backbones to the following issues: 
1) The domain gap between natural images and autonomous driving scenes.
Backbones pre-trained on general 2D recognition tasks fall short of perceiving 3D scenes, especially estimating depth. 
2) The complex structure of current BEV detectors. 
Take BEVFormer~\cite{bevformer} as an example. The supervision signals of 3D bounding boxes and object class labels are separated from the image backbone by the view encoder and the object decoder, each of which is comprised of multiple layers of transformers. 
The gradient flow for adapting general 2D image backbones for autonomous driving tasks is distorted by the stacked transformer layers.

In order to combat the difficulties mentioned above in adapting modern image backbones for BEV recognition, we introduce perspective supervision into BEVFormer, \ie extra supervision signals from perspective-view tasks and directly applied to the backbone. 
It guides the backbone to learn 3D knowledge missing in 2D recognition tasks and overcomes the complexity of BEV detectors, greatly facilitating the optimization of the model.
Specifically, we build a perspective 3D detection head \cite{DD3D} upon the backbone, which takes image features as input and directly predicts the 3D bounding boxes and class labels of target objects. 
The loss of this perspective head, denoted as perspective loss, is added to the original loss (BEV loss) deriving from the BEV head as an auxiliary detection loss. 
The two detection heads are jointly trained with their corresponding loss terms. 
Furthermore, we find it natural to combine the two detection heads into a two-stage BEV detector,  \textbf{BEVFormer v2}. 
Since the perspective head is full-fledged, it could generate high-quality object proposals in the perspective view, which we use as first-stage proposals. 
We encode them into object queries and gather them with the learnable ones in the original BEVFormer, forming hybrid object queries, which are then fed into the second-stage detection head to generate the final predictions.

We conduct extensive experiments to confirm the effectiveness and necessity of our proposed perspective supervision. 
The perspective loss facilitates the adaptation of the image backbone, resulting in improved detection performance and faster model convergence. 
While without this supervision, the model cannot achieve comparable results even if trained with a longer schedule. 
Consequently, we successfully adapt modern image backbones to the BEV model, achieving  63.4\% NDS on nuScenes \cite{nuscenes} test-set.  

Our contributions can be summarized as follows:
 \begin{itemize}
     \item We point out that perspective supervision is key to adapting general 2D image backbones to the BEV model. We add this supervision explicitly by a detection loss in the perspective view. 
     \item We present a novel two-stage BEV detector, BEVFormer v2. It consists of a perspective 3D and a BEV detection head, and the proposals of the former are combined with the object queries of the latter.
     \item We highlight the effectiveness of our approach by combining it with the latest developed image backbones and achieving significant improvements over previous state-of-the-art results on the nuScenes dataset.
 \end{itemize}
\section{Related Works}
\subsection{BEV 3D Object Detector}
Bird's-eye-view (BEV) object detection has attracted more attention recently~\cite{OFT,VPN,pseudo-lidar,LSS,bevformer,CVT,PETR} due to its vast success in autonomous driving systems. 

Early works including OFT~\cite{OFT}, Pseduo LiDAR~\cite{pseudo-lidar}, and VPN~\cite{VPN} shed light on how to transform perspective features into BEV features but either for a single camera or on less well-known tasks. 
OFT~\cite{OFT} pioneered to adopt transformation from 2D image features to 3D BEV features for monocular 3D object detection.
Pseudo LiDAR~\cite{pseudo-lidar}, as its name suggested, created pseudo point clouds through monocular depth estimation and camera intrinsics and processed them in the BEV space subsequently.
VPN~\cite{VPN} was the first to fuse multi-view camera inputs into a top-down view feature map for semantic segmentation.

Modern approaches enjoyed the convenience of integrating features from different perspective view sensors provided by 2D-3D view transformation.
LSS~\cite{LSS} extended OFT by introducing a latent depth distribution during the pooling of BEV pillar features. 
Moreover, LSS pooled over six surrounding images compared with a single in OFT.
Different from the 2D-to-3D lifting in LSS or the 3D-to-2D projection in OFT, CVT~\cite{CVT} utilized camera-aware positional encoding and dense cross attention to bridge perspective-view and BEV-view features.
PETR~\cite{PETR} devised an approach without explicit BEV feature construction. 
Perspective feature maps are element-wisely fused with 3D positional embedding feature maps, and a subsequent DETR-style decoder is applied for object detection.
BEVFormer~\cite{bevformer} leveraged spatial cross-attention for view transformation and temporal self-attention for temporal feature fusion. 
The fully transformer-based structure of BEVFormer makes its BEV features more versatile than other methods, easily supporting non-uniform and non-regular sampling grids.
Besides, as shown in SimpleBEV~\cite{simplebev}, multi-scale deformable attention~\cite{deformable-detr} excels in all lifting strategies.
So we choose to build our detector based on BEVFormer to exploit the strengths mentioned before.

Besides published works, there are many concurrent works due to the popularity of this field.
BEVDet~\cite{BEVDet} introduced rich image-level and BEV-level augmentations for training. 
BEVStereo~\cite{bevstereo} and STS~\cite{STS} both adopted a temporal stereo paradigm for better depth estimation.
PolarFormer~\cite{polarformer} came up with a non-cartesian 3D grid setting.
SimpleBEV~\cite{simplebev} compared different 2D-3D lifting methods.

Unlike existing works that mainly explore the designs for detectors, we focus on adapting modern image backbones into BEV recognition models. 

\begin{figure*}[t]
    \centering
    \includegraphics[width=0.9\textwidth]{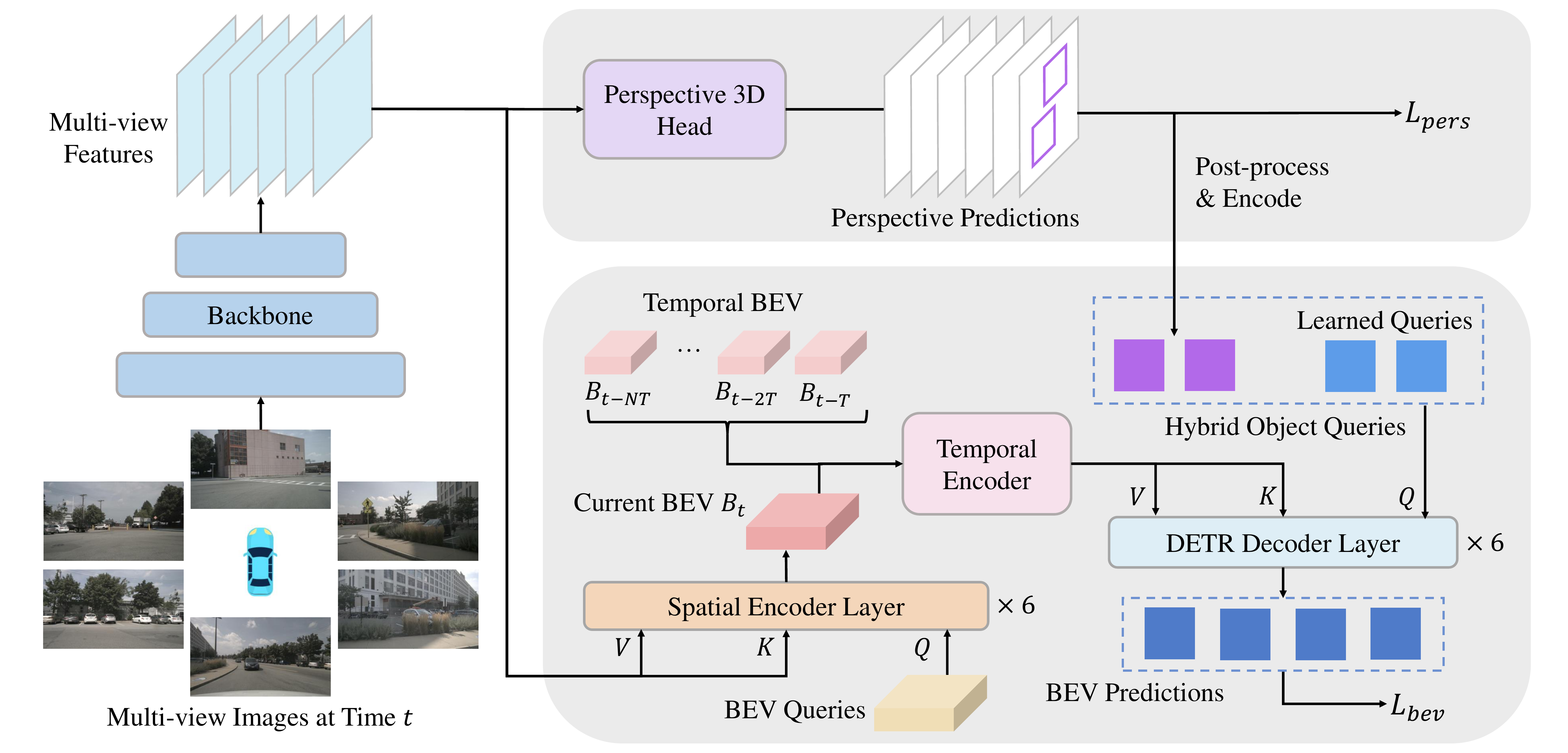}
    \caption{Overall architecture of BEVFormer v2. The image backbone generates features of multi-view images. The perspective 3D head makes perspective predictions which are then encoded as object queries. The BEV head is of encoder-decoder structure. The spatial encoder generates BEV features by aggregating multi-view image features, followed by the temporal encoder that collects history BEV features. The decoder takes hybrid object queries as input and makes the final BEV predictions based on the BEV features. The whole model is trained with the two loss terms of the two detection heads, $L_{pers}$ and $L_{bev}$.}
    \label{fig:architecture}
\end{figure*}

\subsection{Auxiliary Loss in Camera 3D Object Detection}
Auxiliary losses are ubiquitous in monocular 3D object detection as most methods~\cite{MonoDIS,monoflex,fcos3d,DD3D,MonoCon,bevdepth,mv-fcos3d} are built upon 2D detectors like RetinaNet~\cite{RetinaNet} and FCOS~\cite{fcos}.
But those auxiliary losses seldom endowed any explicit meaning for 2D supervisions.
MonoCon~\cite{MonoCon} made the most out of 2D auxiliary by utilizing up to 5 different 2D supervisions.
As for BEV detectors, BEVDepth~\cite{bevdepth} utilized LiDAR point clouds to supervise its intermediate depth network.
MV-FCOS3D++~\cite{mv-fcos3d} introduced perspective supervision for training its image backbone, but the detector itself was supervised by BEV losses alone.
SimMOD~\cite{SimMOD} used 2D auxiliary losses for its monocular proposal head.

Different from previous methods, our method adopted an end-to-end perspective supervision approach without using extra data such as LiDAR point clouds.


\subsection{Two-stage 3D Object Detector}
Although two-stage detectors are common in LiDAR-based 3D object detection~\cite{MV3D,AVOD,centerpoint,lidar-rcnn,frustum-pointnet,transfusion,SimMOD}, their application in camera-based 3D detection is far less well known. 
MonoDIS~\cite{MonoDIS} used RoIAlign to extract image features from 2D boxes and to regress 3D boxes subsequently.
SimMOD~\cite{SimMOD} employed a monocular 3D head for making proposals and a DETR3D \cite{DETR3D} head for the final detection.
However, using the same features from the perspective backbone in both stages provides no information gain for the second-stage head. 
We suppose that this is the main reason why two-stage detectors were far less popular in camera-based 3D detection. 
Instead, our two-stage detector utilizes features from both perspective and BEV view and thus enjoys information in both image and BEV space.

\section{BEVFormer v2}
Adapting modern 2D image backbones for BEV recognition without cumbersome depth pre-training could unlock many possibilities for downstream autonomous driving tasks. 
In this work, we propose BEVformer v2, a two-stage BEV detector that incorporates both BEV and perspective supervision for a hassle-free adoption of image backbones in BEV detection.

\subsection{Overall Architecture} As illustrated in Fig. \ref{fig:architecture}, BEVFormer v2 mainly consists of five components: an image backbone, a perspective 3D detection head, a spatial encoder, a revamped temporal encoder, and a BEV detection head. 
Compared with the original BEVFormer~\cite{bevformer}, changes are made for all components except the spatial encoder.
Specifically, all image backbones used in BEVFormer v2 are not pre-trained with any autonomous driving datasets or depth estimation datasets. 
A perspective 3D detection head is introduced to facilitate the adaptation of 2D image backbones and generate object proposals for the BEV detection head.
A new temporal BEV encoder is adopted for better incorporating long-term temporal information.
The BEV detection head now accepts a hybrid set of object queries as inputs. We combine the first-stage proposals and the learned object queries to
form the new hybrid object queries for the second stage.

\subsection{Perspective Supervision}
\begin{figure}[t]
    \centering
    \includegraphics[width=1.0\linewidth]{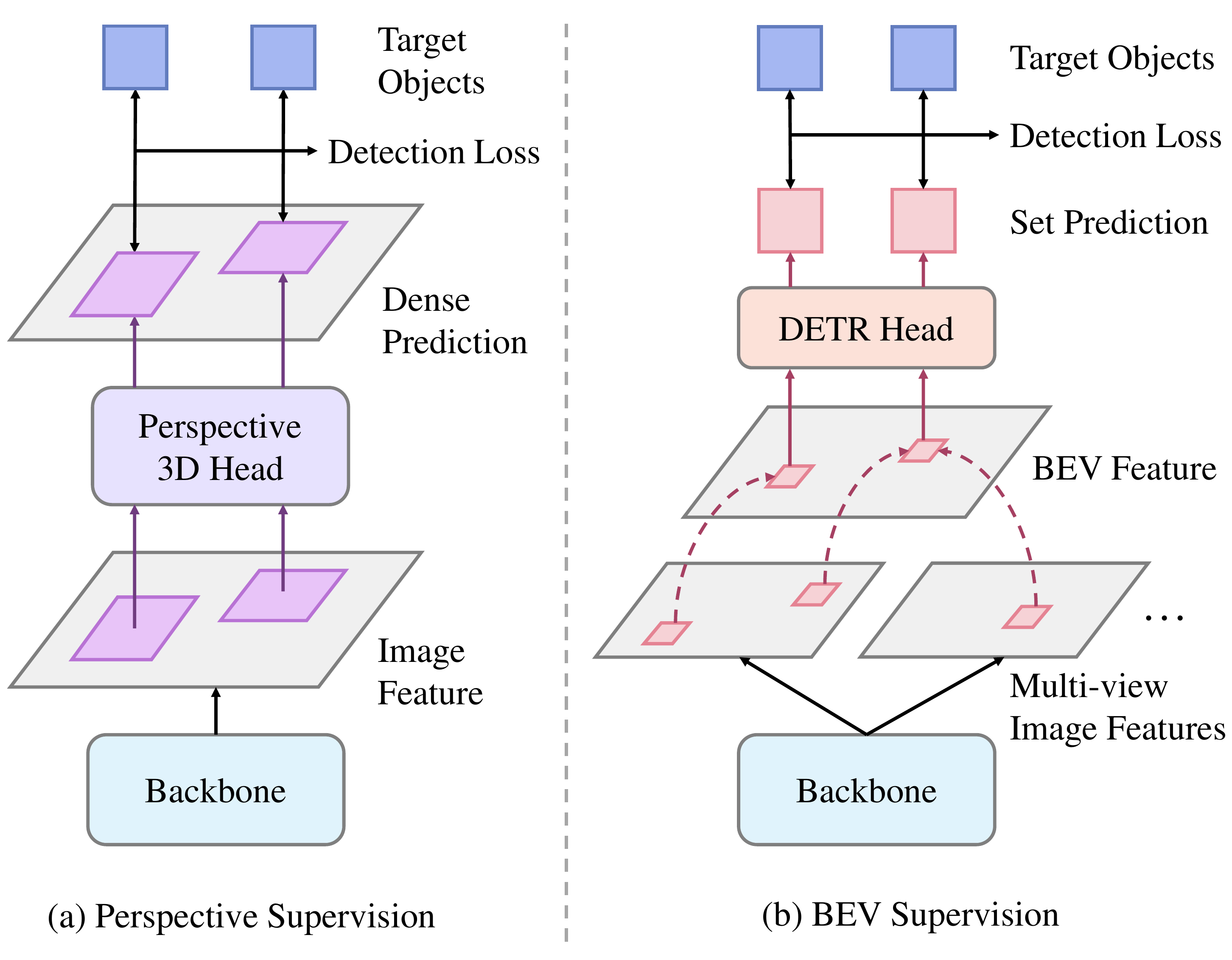}
    \caption{Comparison of perspective supervision (a) and BEV supervision (B). The supervision signals of the perspective detector are dense and direct to the image feature, while those of the BEV detector are sparse and indirect.}
    \label{fig:supervision}
\end{figure}

We first analyze the problem of the bird's-eye-view models to explain why additional supervision is necessary. 
A typical BEV model maintains grid-shaped features attached to the BEV plane, where each grid aggregates 3D information from the features at corresponding 2D pixels of multi-view images. 
It predicts the 3D bounding boxes of the target objects based on the BEV features, and we name this supervision imposed on BEV features as BEV supervision. 
Take BEVformer \cite{bevformer} as an example, it uses an encoder-decoder structure to generate and exploit the BEV features. 
The encoder assigns each grid cell on the BEV plane with a set of 3D reference points and projects them onto multi-view images as 2D reference points. 
After that, it samples image features around 2D reference points and utilizes spatial cross-attention to aggregate them into the BEV features. 
The decoder is a Deformable DETR \cite{deformable-detr} head that predicts 3D bounding boxes in the BEV coordinate with a small fixed number of object queries. 
Fig. \ref{fig:supervision} shows the two underlying issues of BEV supervision introduced by the 3D-to-2D view transformation and the DETR \cite{DETR} head: 
\begin{itemize}
    \item The supervision is implicit with respect to the image features.
    The loss is directly applied to the BEV features, while it becomes indirect after 3D-to-2D projection and attentive sampling of the image features.
    \item The supervision is sparse to the image features. 
    Only a small number of BEV grids attended by the object queries contribute to the loss. Consequently, only sparse pixels around the 2D reference points of those grids obtain the supervisory signal.
\end{itemize}
Therefore, inconsistency emerges during training that the BEV detection head relies on the 3D information contained in the image features, but it provides insufficient guidance for the backbone on how to encode such information.

\label{sec:perspective_supervision}

Previous BEV methods do not severely suffer from this inconsistency, and they may not even realize this problem. 
This is because their backbones either have relatively small scales or have been pre-trained on 3D detection tasks with a monocular detection head. 
In contrast to the BEV head, the perspective 3D head makes per-pixel predictions upon the image features, offering much richer supervision signals for adapting 2D image backbones. 
We define this supervision imposed on the image feature as perspective supervision. 
As shown in Fig. \ref{fig:supervision}, different from the BEV supervision, the perspective detection loss is directly and densely applied to the image features. 
We suppose that perspective supervision explicitly guides the backbone to perceive 3D scenes and extract useful information, e.g., the depths and orientations of the objects, overcoming the drawbacks of BEV supervision, thus is essential when training BEV models with modern image backbones.

\subsection{Perspective Loss}
As analyzed in the previous session, perspective supervision is the key to optimizing BEV models. 
In BEVformer v2, we introduce perspective supervision via an auxiliary perspective loss. 
Specifically, a perspective 3D detection head is built upon the backbone to detect target objects in the perspective view. 
We adopt an FCOS3D~\cite{fcos3d}-like detection head, which predicts the center location, size, orientation, and projected center-ness of the 3D bounding boxes. 
The detection loss of this head, denoted as perspective loss $\mathcal{L}_{pers}$, serves as the complement to the BEV loss $\mathcal{L}_{bev}$, facilitating the optimization of the backbone. 
The whole model is trained with a total objective
\begin{equation}
\mathcal{L}_{total} = \lambda_{bev}\mathcal{L}_{bev}+\lambda_{pers}\mathcal{L}_{pers}.
\end{equation}

\subsection{Ravamped Temporal Encoder}
BEVFormer uses recurrent temporal self-attention for incorporating historical BEV features.
But the temporal encoder falls short of utilizing long-term temporal information, simply increasing the recurrent steps from 4 to 16 yields no extra performance gain.

We redesign the temporal encoder for BEVFormer v2 by a using simple warp and concatenate strategy.
Given a BEV feature $B_k$ at a different frame $k$, we first bi-linearly warp $B_k$ into the current frame as $B_k^t$ according to the reference frame transformation matrix $T_k^{t} = [\bf{R} | \bf{t}] \in \text{SE3}$ between frame $t$ and frame $k$. 
We then concatenate previous BEV features with the current BEV feature along the channel dimension and employ residual blocks for dimension reduction.
To maintain a similar computation complexity as the original design, we use the same number of historical BEV features but increase the sampling interval.
Besides benefiting from long-term temporal information, the new temporal encoder also unlocks the possibility of utilizing future BEV features in the offline 3D detection setting.

\subsection{Two-stage BEV Detector}
Though jointly training two detection heads has provided enough supervision, we obtain two sets of detection results separately from different views. 
Rather than take the predictions of the BEV head and discard those of the perspective head or heuristically combine two sets of predictions via NMS, we design a novel structure that integrates the two heads into a two-stage predicting pipeline, namely, a two-stage BEV detector. 
The object decoder in the BEV head, a DETR \cite{DETR} decoder, uses a set of learned embeddings as object queries, which learns where the target objects possibly locate through training. 
However, randomly initialized embeddings take a long time to learn appropriate positions. 
Besides, learned object queries are fixed for all images during inference, which may not be accurate enough since the spatial distribution of objects may vary. 
To address these issues, the predictions of the perspective head are filtered by post-processing and then fused into the object queries of the decoder, forming a two-stage process. 
These hybrid object queries provide candidate positions with high scores (probability), making it easier for the BEV head to capture target objects in the second stage. The details of the decoder with hybrid object queries will be described later. 
It should be noticed that the first-stage proposals are not necessarily from a perspective detector, e.g., from another BEV detector, but experiments show that only the predictions from the perspective view are helpful for the second-stage BEV head. 

\subsection{Decoder with Hybrid Object Queries}
\begin{figure}[t]
    \centering
    \includegraphics[width=1.0\linewidth]{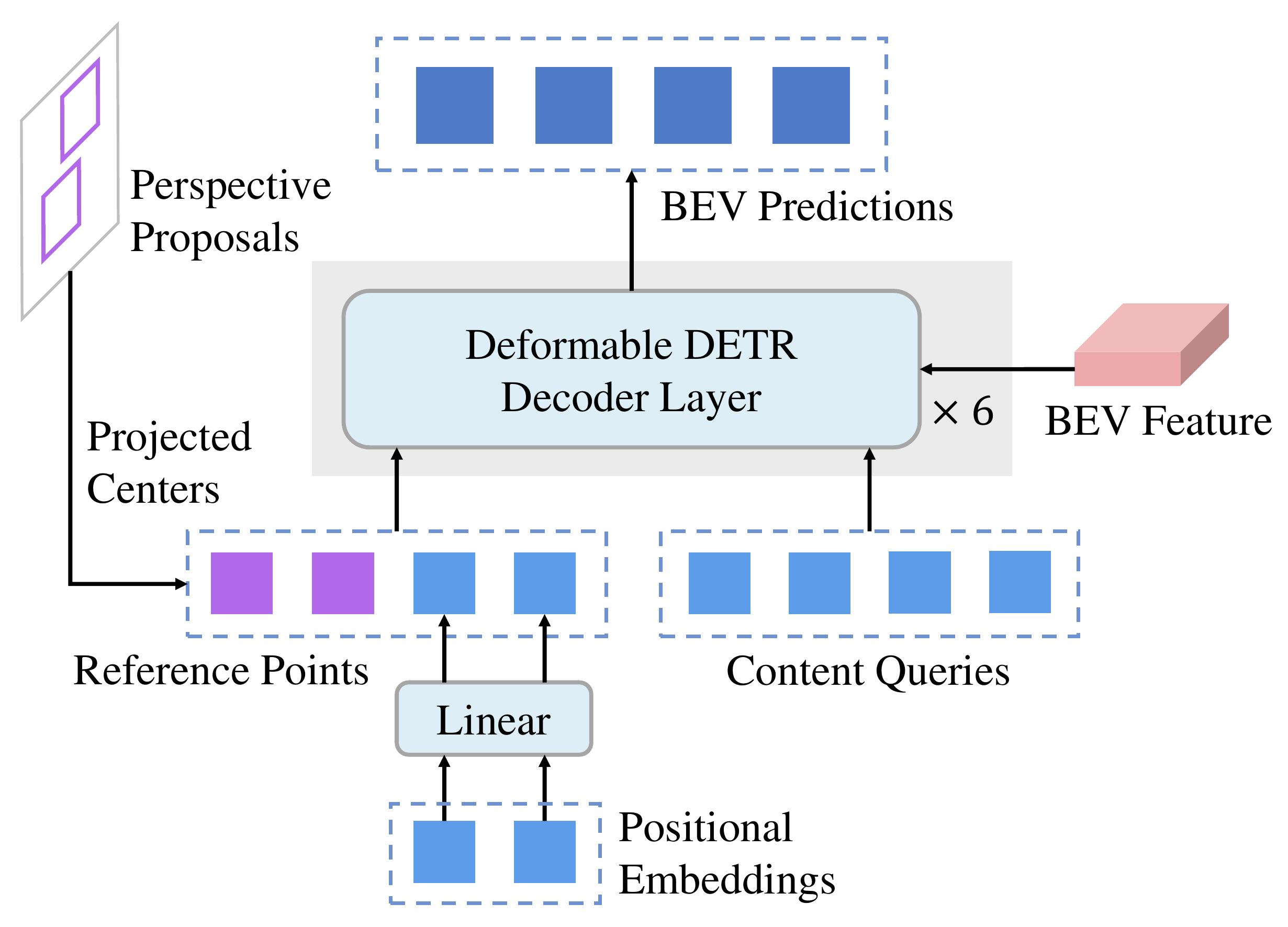}
    \caption{The decoder of the BEV head in BEVFromer v2. The projected centers of the first-stage proposals are used as per-image reference points (purple ones), and they are combined with per-dataset learnded content queries and positional embeddings (blue ones) as hybrid object queries.}
    \label{fig:decoder}
    \vspace{-0.3cm}
\end{figure}

To fuse the first-stage proposals into the object queries of the second stage, the decoder of the BEV head in BEVformer v2 is modified based on the Deformable DETR~\cite{deformable-detr} decoder used in BEVFormer \cite{bevformer}. 
The decoder consists of stacked alternated self-attention and cross-attention layers. 
The cross-attention layer is a deformable attention module \cite{deformable-detr} that takes the following three elements as input. 
(1) Content queries, the query features to produce sampling offsets and attention weights.
(2) Reference points, the 2D points on the value feature as the sampling reference of each query.
(3) Value features, the BEV feature to be attended.
In the original BEVFormer \cite{bevformer}, the content queries are a set of learned embeddings and reference points are predicted with a linear layer from a set of learned positional embeddings. 
In BEVformer v2, we obtain proposals from the perspective head and select a part of them via post-processing.
As illustrated in Fig. \ref{fig:decoder}, the projected box centers on the BEV plain of the selected proposals are used as per-image reference points and are combined with the per-dataset ones generated from positional embeddings. 
The per-image reference points directly indicate the possible positions of objects on the BEV plain, making it easier for the decoder to detect target objects. 
However, a small part of objects may not be detected by the perspective head due to occlusion or appearing at the boundary of two adjacent views.
To avoid missing these objects, we also keep the original per-dataset reference points to capture them by learning a spatial prior. 

\section{Experiments}
\subsection{Dataset and Metrics.} 
The nuScenes 3D detection benchmark \cite{nuscenes} consists of 1000 multi-modal videos of roughly 20s duration each, and the key samples are annotated at 2Hz. 
Each sample consists of images from 6 cameras covering the full 360-degree field of view. 
The videos are split into 700 for training, 150 for validation, and 150 for testing. The detection task contains 1.4M annotated 3D bounding boxes of 10 object classes. 
The nuScenes computes the mean average precision (mAP) over four different thresholds using center distance on the ground plane, and it contains five true-positive metrics, namely, ATE, ASE, AOE, AVE, and AAE, for measuring translation, scale, orientation, velocity, and attribute errors, respectively. 
In addition, it also defines a nuScenes detection score (NDS) by combining the detection accuracy (mAP) with the five true-positive metrics.

\setlength{\tabcolsep}{3pt}
\setlength{\doublerulesep}{2\arrayrulewidth}
\renewcommand{\arraystretch}{1.1}
\begin{table*}[t]
    \caption{3D detection results on the nuScenes $test$ set of BEVFormer v2 and other SoTA methods.$^\dagger$ indicates that V2-99 \cite{Vovnet} was pre-trained on the depth estimation task with extra data \cite{DD3D}. $^\ddagger$ indicates methods with CBGS which will elongate 1 epoch into 4.5 epochs. We choose to only train BEVFormer v2 for 24 epochs to compare fairly with previous methods.}
    \label{table:sota_nuscenes}
    \centering

    \begin{tabular}{l|c|c|c|cc|ccccc}
        \toprule
        Method & Backbone & Epoch & Image Size & NDS & mAP & mATE & mASE & mAOE & mAVE & mAAE \\ 
        \midrule
        BEVFormer~\cite{bevformer} & V2-99$^\dagger$ & 24 & 900 $\times$ 1600 & 0.569 & 0.481 & 0.582 & 0.256 & 0.375 & 0.378 & 0.126 \\
        PolarFormer~\cite{polarformer} & V2-99$^\dagger$ & 24 & 900 $\times$ 1600 & 0.572 & 0.493 & 0.556 & 0.256 & 0.364 & 0.440 & 0.127 \\
        PETRv2~\cite{petrv2} & GLOM & 24 & 640 $\times$ 1600 & 0.582 & 0.490 & 0.561 & 0.243 & 0.361 & 0.343 & 0.120 \\
        BEVDepth~\cite{bevdepth} & V2-99$^\dagger$ & 90$^\ddagger$ & 640 $\times$ 1600 & 0.600 & 0.503 & 0.445 & 0.245 & 0.378 & 0.320 & 0.126 \\
        BEVStereo~\cite{bevstereo} & V2-99$^\dagger$ & 90$^\ddagger$ & 640 $\times$ 1600 & 0.610 & 0.525 & 0.431 & 0.246 & 0.358 & 0.357 & 0.138 \\
        \midrule
        \textbf{BEVFormer v2} & InternImage-B & 24 & 640 $\times$ 1600 & 0.620 & 0.540 & 0.488 & 0.251 & 0.335 & 0.302 & 0.122 \\
        \textbf{BEVFormer v2} & InternImage-XL & 24 & 640 $\times$ 1600 & 0.634 & 0.556 & 0.456 & 0.248 & 0.317 & 0.293 & 0.123 \\
        \bottomrule
    \end{tabular}

\end{table*}
\setlength{\tabcolsep}{3pt}
\setlength{\doublerulesep}{2\arrayrulewidth}
\renewcommand{\arraystretch}{1.1}
\begin{table*}[t]
    \caption{The detection results of 3D detectors with different combinations of view supervision on the nuScenes $val$ set. All models are trained without temporal information.}
    \label{table:view_supervision}
    \centering

    \begin{tabular}{c|c|c|cc|cccccc}
        \toprule
        View Supervision & Backbone & Epoch & NDS & mAP & mATE & mASE & mAOE & mAVE & mAAE   \\ 
        \midrule 
        Perspective Only & ResNet-101 & 48 & 0.412 & 0.323 & 0.737 & 0.268 & 0.377 & 0.943 & 0.167 \\ 
        BEV Only & ResNet-101 & 48 & 0.426 & 0.355 & 0.751 & 0.275 & 0.429 & 0.847 & 0.215 \\ 
        Perspective \& BEV & ResNet-101 & 48 & 0.451 & 0.374 & 0.730 & 0.270 & 0.379 & 0.773 & 0.205 \\ 
        BEV \& BEV & ResNet-101 & 48 & 0.428 & 0.350 & 0.750 & 0.279 & 0.388 & 0.842 & 0.210 \\ 
        \midrule
        \bottomrule
    \end{tabular}
    
\end{table*}

\subsection{Experimental Settings}
We conduct experiments with multiple types of backbones: ResNet~\cite{ResNet}, DLA~\cite{DLA}, VoVNet~\cite{Vovnet}, and InternImage \cite{InternImage}. 
All the backbones are initialized with the checkpoints pre-trained on the 2D detection task of the COCO dataset \cite{COCO}. 
Except for our modification, we follow the default settings of BEVFormer~\cite{bevformer} to construct the BEV detection head. 
In Tab. \ref{table:sota_nuscenes} and Tab. \ref{table:trick}, the BEV head utilizes temporal information with the new temporal encoder. For other experiments, we employ the single-frame version that only uses the current frame, like BEVFormer-S~\cite{bevformer}. 
For the perspective 3D detection head, we adopt the implementation in DD3D~\cite{DD3D} with camera-aware depth parameterization. 
The loss weight of perspective loss and BEV loss are set as $\lambda_{bev}=\lambda_{pers}=1$. 
We use AdamW~\cite{AdamW} optimizer and set the base learning rate as 4e-4. 

\subsection{Benchmark Results}
We compare our proposed BEVFormer v2 with existing state-of-the-art BEV detectors including BEVFormer~\cite{bevformer}, PolarFormer~\cite{polarformer}, PETRv2~\cite{petrv2}, BEVDepth~\cite{bevdepth}, and BEVStereo~\cite{bevstereo}. 
We report the 3D object detection results on the nuScenes test set in Tab. \ref{table:sota_nuscenes}. 
The V2-99 \cite{Vovnet} backbone used by BEVFormer, PolarFormer, BEVDepth, and BEVStereo have been pre-trained on the depth estimation task with extra data and then fine-tuned by DD3D \cite{DD3D} on the nuScenes dataset \cite{nuscenes}.
On the contrary, the InternImage \cite{InternImage} backbone we employ is initialized with the checkpoint from COCO~\cite{COCO} detection task without any 3D pretraining.
InternImage-B has a similar number of parameters to V2-99, but better reflects the progress of modern image backbone design.
We can observe that BEVFormer v2 with InternImage-B backbone outperforms all existing methods, showing that with the perspective supervision, backbones pre-trained on monocular 3D tasks are no longer necessary.  
BEVFormer v2 with InternImage-XL outperforms all entries on the nuScenes camera 3D objection leaderboard with 63.4\% NDS and 55.6\% mAP, surpassing the second-place method BEVStereo by 2.4\% NDS and 3.1\% mAP. 
This significant improvement reveals the huge benefit of unleashing the power of modern image backbone for BEV recognition.

\setlength{\tabcolsep}{3pt}
\setlength{\doublerulesep}{2\arrayrulewidth}
\renewcommand{\arraystretch}{1.1}
\begin{table*}[ht]
    \caption{The results of perspective supervision with different 2D image backbones on the nuScenes $val$ set. `BEV Only' and `Perspective \& BEV' are the same as Tab. \ref{table:view_supervision}. All the backbones are initialized with COCO~\cite{COCO} pretrained weights and all models are trained without temporal information.}
    \label{table:large_backbone}
    \centering

    \begin{tabular}{c|c|c|cc|cccccc}
        \toprule
        Backbone & Epoch & View Supervision & NDS & mAP & mATE & mASE & mAOE & mAVE & mAAE   \\ 
        \midrule
        ResNet-50 & 48 & BEV Only & 0.400 & 0.327 & 0.795 & 0.277 & 0.479 &0.871 & 0.210 \\ 
        ResNet-50 & 48 & Perspective \& BEV & 0.428 & 0.349 & 0.750 & 0.276 & 0.424 & 0.817 & 0.193 \\
        \midrule
        DLA-34 & 48 & BEV Only & 0.403 & 0.338 & 0.772 & 0.279 & 0.483 & 0.919 & 0.206 \\ 
        DLA-34 & 48 & Perspective \& BEV & 0.435 & 0.358 & 0.742 & 0.274 & 0.431 & 0.801 & 0.186 \\  
  
        \midrule
        ResNet-101 & 48 & BEV Only & 0.426 & 0.355 & 0.751 & 0.275 & 0.429 & 0.847 & 0.215  \\ 
        ResNet-101 & 48 & Perspective \& BEV & 0.451 & 0.374 & 0.730 & 0.270 & 0.379 & 0.773 & 0.205  \\   
        \midrule
        VoVNet-99 & 48 & BEV Only & 0.441 & 0.367  & 0.734 & 0.271 & 0.402 & 0.815  & 0.205 \\ 
        VoVNet-99 & 48 & Perspective \& BEV & 0.467 & 0.396  & 0.709  & 0.274 & 0.368 & 0.768 & 0.196 \\ 
        \midrule
        InternImage-B & 48 & BEV Only & 0.455 & 0.398  & 0.712 & 0.283 & 0.411 & 0.826  & 0.204 \\ 
        InternImage-B & 48 & Perspective \& BEV & 0.485 & 0.417  & 0.696  & 0.275 & 0.354 & 0.734 & 0.182 \\ 
        \bottomrule
    \end{tabular}
    
\end{table*}
\setlength{\tabcolsep}{3pt}
\setlength{\doublerulesep}{2\arrayrulewidth}
\renewcommand{\arraystretch}{1.1}
\begin{table*}[t]
    \caption{Comparing models with BEV supervision only and with both Perspective \& BEV supervision under different training epochs. The models are evaluated on the nuScenes $val$ set. All models are trained without temporal information.}
    \label{table:long_schedule}
    \centering

    \begin{tabular}{c|c|c|cc|cccccc}
        \toprule
        View Supervision & Backbone & Epoch  & NDS & mAP & mATE & mASE & mAOE & mAVE & mAAE   \\ 
        \midrule 
        \multirow{3}{*}{BEV Only} & \multirow{3}{*}{ResNet-50} & 24 & 0.379 & 0.322 & 0.803 & 0.280 & 0.549 & 0.954 & 0.240 \\ 
         &  & 48 & 0.400 & 0.327 & 0.795 & 0.277 & 0.479 & 0.871 & 0.210 \\ 
         &  & 72 & 0.410 & 0.335 & 0.771 & 0.280 & 0.458 & 0.848 & 0.216 \\ 
        \midrule
        \multirow{3}{*}{Perspective \& BEV} & \multirow{3}{*}{ResNet-50} & 24 &  0.414 & 0.351 & 0.732 & 0.271 & 0.505 & 0.899 & 0.204 \\ 
         &  & 48 & 0.428 & 0.349 & 0.750 & 0.276 & 0.424 & 0.817 & 0.193 \\ 
         &  & 72 & 0.428 & 0.351 & 0.741 & 0.279 & 0.419 & 0.835 & 0.196 \\ 
        \midrule
        \bottomrule
    \end{tabular}
\end{table*}

\subsection{Ablations and Analyses}
\subsubsection{Effectiveness of Perspective Supervision} 
To confirm the effectiveness of perspective supervision, we compare 3D detectors with different view supervision combinations in Tab. \ref{table:view_supervision}, including 
(1) Perspective \& BEV, the proposed BEVFormer v2, a two-stage detector integrating a perspective head and a BEV head. 
(2) Perspective Only, the single-stage perspective detector in our model. 
(3) BEV Only, the single-stage BEV detector in our model without hybrid object queries. 
(4) BEV \& BEV, a two-stage detector with two BEV heads, \ie, replace the perspective head in our model with another BEV head that utilizes BEV features to make proposals for the hybrid object queries. 

Compared with the Perspective Only detector, the BEV Only detector achieves better NDS and mAP by leveraging multi-view images, but its mATE and mAOE are higher, indicating the underlying issues of BEV supervision. 
Our Perspective \& BEV detector achieves the best performance and outperforms BEV Only detector with a margin of 2.5\% NDS and 1.9\% mAP. 
Specifically, the mATE, mAOE, and mAVE of Perspective \& BEV detector are significantly lower than those of BEV Only detector. 
This remarkable improvement mainly from the following two aspects: 
(1) Backbones pre-trained on normal vision tasks cannot capture some properties of objects in 3D scenes, including depth, orientation, and velocity, while backbones guided by perspective supervision are capable of extracting information about such properties.
(2) Compared to a fixed set of object queries, our hybrid object queries contain the first-stage predictions as reference points, helping the BEV head to locate target objects.
To further ensure that the improvement is not brought by the two-stage pipeline, we introduce the BEV \& BEV detector for comparison. 
It turns out that BEV \& BEV is on par with BEV Only and is not comparable with Perspective \& BEV. 
Therefore, only constructing the first-stage head and applying auxiliary supervision in the perspective view is helpful for BEV models.

\subsubsection{Generalization of Perspective Supervision}
The proposed perspective supervision is expected to benefit backbones of different architectures and sizes. 
We construct BEVFormer v2 on a series of backbones commonly used for 3D object detection tasks: ResNet \cite{ResNet}, DLA \cite{DLA}, VoVNet \cite{Vovnet}, and InternImage \cite{InternImage}. The results are reported in Tab. \ref{table:large_backbone}. 
Compared to pure BEV detector, BEVForemr v2 (BEV \& perspective) boosts NDS by around 3\% and mAP by around 2\% for all the backbones, manifesting that it generalizes to different architectures and model sizes. 
We suppose that the additional perspective supervision can be a general scheme for training BEV models, especially when adapting large-scale image backbones without any 3D pre-training.

\setlength{\tabcolsep}{3pt}
\setlength{\doublerulesep}{2\arrayrulewidth}
\renewcommand{\arraystretch}{1.1}
\begin{table*}[ht]
    \caption{Comparison of different choices for the perspective head and the BEV head in BEVFormer v2. The models are evaluated on the nuScenes $val$ set. All models are trained without temporal information.}
    \label{table:detector_choice_for_two_view}
    \centering
    \vspace{-5pt}
    \begin{tabular}{cc|c|c|cc|ccccc}
        \toprule
        Perspective View & BEV View & Backbone & Epoch & NDS & mAP & mATE & mASE & mAOE & mAVE & mAAE   \\ 
        \midrule 
        DD3D & Deformable DETR & ResNet-50 & 48 & 0.428 & 0.349 & 0.750 & 0.276 & 0.424 & 0.817 & 0.193 \\ 
        DD3D & Group DETR & ResNet-50 & 48 & 0.445 & 0.353 & 0.725 & 0.276 & 0.366 & 0.767 & 0.180 \\ 
         DETR3D & Deformable DETR & ResNet-50 & 48 & 0.409 & 0.335 & 0.765 & 0.276 & 0.469 & 0.877 & 0.198 \\ 
        DETR3D & Group DETR & ResNet-50 & 48 & 0.423 & 0.351 & 0.743 & 0.279 & 0.466 & 0.844 & 0.201 \\ 
        \midrule
        \bottomrule
    \end{tabular}
    
\end{table*}
\let\ck\checkmark

\setlength{\tabcolsep}{3pt}
\setlength{\doublerulesep}{2\arrayrulewidth}
\renewcommand{\arraystretch}{1.0}
\begin{table*}[t]
    \caption{Ablation study of bells and whistles of BEVFormer v2 on the nuScenes $val$ set. All models are trained with a ResNet-50 backbone and temporal information. `Pers', `IDA', `Long', and `Bi' denotes perspective supervision, image-level data augmentation, long temporal interval, and bi-directional temporal encoder, respectively.}
    \label{table:trick}
    \centering
    \vspace{-5pt}
    \begin{tabular}{l|c|cccc|cc|cc|cccccc}
        \toprule
        Method & Epoch & Pers & IDA & Long & Bi & NDS & mAP & mATE & mASE & mAOE & mAVE & mAAE   \\ 
        \midrule
        Baseline & 24 & \ck & & & & 0.478 & 0.368 & 0.709 & 0.282 & 0.452 & 0.427 & 0.191 \\ 
        Image-level Data Agumentation & 24 & \ck & \ck & & & 0.489 & 0.386 & 0.690 & 0.273 & 0.482 & 0.395 & 0.199 \\
        Longer Temporal Interval & 24 & \ck & \ck & \ck & & 0.498 & 0.388 & 0.679 & 0.276 & 0.417 & 0.403 & 0.189 \\
        Bi-directional Temporal Encoder & 24 & \ck & \ck & \ck & \ck & 0.529 & 0.423 & 0.618 & 0.273 & 0.413 & 0.333 & 0.181 \\
        All but Perspective & 24 & & \ck & \ck & \ck & 0.507 & 0.397 & 0.636 & 0.281 & 0.455 & 0.356 & 0.190 \\
        \bottomrule
    \end{tabular}
    
\end{table*}

\subsubsection{Choice of Training Epochs}
We train the BEV Only model and our BEVFormer v2 (BEV \& Perspective) for different epochs to see how many the two models take to achieve convergence.
Tab. \ref{table:long_schedule} shows that our BEV \& Perspective model converges faster than the BEV Only one, confirming that auxiliary perspective loss facilitates the optimization.
The BEV Only model obtains marginal improvement if it is trained for more time. 
But the gap between the two models remains at 72 epochs and may not be eliminated even for longer training, which indicates that the image backbones cannot be well adapted by BEV supervision alone.
According to Tab. \ref{table:long_schedule}, training for 48 epochs is enough for our model, and we keep this fixed for other experiments unless otherwise specified. 

\subsubsection{Choice of Detection Heads}
Various types of perspective and BEV detection heads can be used in our BEVFormer v2. 
We explore several representative methods to choose the best for our model: for the perspective head, the candidates are DD3D \cite{DD3D} and DETR3D \cite{DETR3D}; for the BEV head, the candidates are Deformable DETR \cite{deformable-detr} and Group DETR \cite{groupdetr}. 
DD3D is a single-stage anchor-free perspective head that makes dense per-pixel predictions upon the image feature. 
DETR3D, on the contrary, uses 3D-to-2D queries to sample image features and to propose sparse set predictions. 
However, according to our definition, it belongs to perspective supervision since it utilizes image features for the final prediction without generating BEV features, \ie, the loss is directly imposed on the image features.
As shown in Tab. \ref{table:detector_choice_for_two_view}, DD3D is better than DETR3D for the perspective head, which supports our analysis in Sec. \ref{sec:perspective_supervision}. 
Dense and direct supervision offered by DD3D is helpful for BEV models, while sparse supervision of DETR3D does not overcome the drawbacks of BEV heads. 
Group DETR head is an extension of Deformable DETR head that utilizes grouped object queries and self-attention within each group.
Group DETR achieves better performance for the BEV head, but it costs more computation.
Therefore, we employ DD3D head and Group DETR head in Tab. \ref{table:sota_nuscenes} and keep the same Deformable DETR head as BEVformer \cite{bevformer} in other ablations.

\subsubsection{Ablations of Bells and Whistles}
\label{sec:bells_and_whistles}
In Tab. \ref{table:trick}, we ablate the bells and whistles employed in our BEVFormer v2 to confirm their contributions to the final result, including 
(1) Image-level data augmentation (IDA). The images are randomly flipped horizontally. 
(2) Longer temporal interval. Rather than use continuous frames with an interval of 0.5 seconds in BEVFormer \cite{bevformer}, our BEVFormer v2 samples history BEV features with an interval of 2 seconds.
(3) Bi-directional Temporal Encoder. For offline 3D detection, the temporal encoder in our BEVFormer v2 can utilize future BEV features.
With longer temporal intervals, our model can gather information from more ego positions at different time stamps, which helps estimate the orientation of the objects and results in a much lower mAOE.
In the offline 3D detection setting, the bi-directional temporal encoder could provide additional information from future frames and improves the performance of the model by a large margin.
We also ablate the perspective supervision in case of applying all bells and whistles.
As shown in Tab. \ref{table:trick}, perspective supervision boosts NDS by 2.2 \% and mAP by 2.6\%, which contributes to the major improvement.

\section{Conclusion}
Existing works have paid much effort into designing and improving the detectors for bird's-eye-view (BEV) recognition models, but they usually get stuck to specific pre-trained backbones without further exploration.
In this paper, we aim to unleash the full power of modern image backbones on BEV models.
We owe the struggle of adapting general 2D image backbones to the optimization problem of the BEV detector. 
To address this issue, we introduce perspective supervision into the BEV model by adding auxiliary loss from an extra perspective 3D detection head.
In addition, we integrate the two detection heads into a two-stage detector, namely, BEVFormer v2. 
The full-fledged perspective head provides first-stage object proposals, which are encoded into object queries of the BEV head for the second-stage prediction.
Extensive experiments verify the effectiveness and generality of our proposed method.
The perspective supervision guides 2D image backbones to perceive 3D scenes of autonomous driving and helps the BEV model achieve faster convergence and better performance, and it is suitable for a wide range of backbones.
Moreover, we successfully adapt large-scale backbones to BEVFormer v2, achieving new SoTA results on the nuScenes dataset. 
We suppose that our work paves the way for future researchers to explore
better image backbone designs for BEV models.

\noindent\textbf{Limitations.} Due to computation and time limitations, we currently do not test our method on more large-scale image backbones. We have finished a preliminary verification of our method on a spectrum of backbones, and we will extend the model sizes in the future.

{\small
\bibliographystyle{ieee_fullname}
\bibliography{egbib}
}

\appendix
\clearpage
\section{Implementation Details}
In this section, we present more implementation details of the proposed method and experiments.

\subsection{Training Settings}
 In \cref{table:training_setting_main}, we provide the hyper-parameters and training recipes of BEVformer v2 used for InternImage-B~\cite{InternImage} and InternImage-XL backbones in \cref{table:sota_nuscenes}.

\let\ck\checkmark

\setlength{\tabcolsep}{5pt}
\setlength{\doublerulesep}{2\arrayrulewidth}
\renewcommand{\arraystretch}{1.0}
\begin{table}[ht]
    \caption{Training settings of BEVformer v2 with InternImage backbones for the main results.}
    \label{table:training_setting_main}
    \centering
    \begin{tabular}{l|cc}
        backbone & InternImage-B & InternImage-XL \\
        \hline
        training epochs & 24 & 24 \\
        batch size & 16 & 32 \\
        optimizer & AdamW & AdamW \\
        base learning rate & 4e-4 & 5e-4 \\
        weight decay & 0.01 & 0.01 \\
        lr schedule & step decay & step decay\\
        layer-wise lr decay & 0.96 & 0.94 \\
        warmup iters & 2000 & 2000  \\
        warmup schedule & linear & linear \\
        gradient clip & 35 & 35 \\
        \hline
        image size & 640 $\times$ 1600 & 640 $\times$ 1600 \\
        IDA & \ck & \ck \\
        temporal interval & 4 seconds & 4 seconds \\
        bi-directional & \ck & \ck \\
        \hline
    \end{tabular}
    
\end{table}

In Tab. \ref{table:large_backbone}, we also construct our BEVFormer v2 detector on other backbones, including ResNet-50~\cite{ResNet}, DLA-34~\cite{DLA}, ResNet-101~\cite{ResNet}, and VoVNet-99~\cite{Vovnet}. We list their training settings in Tab. \ref{table:training_setting_other}.

\let\ck\checkmark

\setlength{\tabcolsep}{5pt}
\setlength{\doublerulesep}{2\arrayrulewidth}
\renewcommand{\arraystretch}{1.0}
\begin{table}[ht]
    \caption{Training settings of BEVformer v2 with other backbones.}
    \label{table:training_setting_other}
    \centering
    \begin{tabular}{l|cccc}
        backbone & R50 & DLA34 & R101 & V2-99 \\
        \hline
        batch size & \multicolumn{4}{c}{16} \\
        optimizer & \multicolumn{4}{c}{AdamW} \\
        base lr & \multicolumn{4}{c}{4e-4} \\
        backbone lr & 2e-4 & 2e-4 & 4e-5 & 4e-5 \\
        weight decay & \multicolumn{4}{c}{0.01} \\
        \hline
    \end{tabular}
    
\end{table}

\subsection{Network Architecture}
In BEVformer v2, the image backbone yields 3 levels of feature maps of stride 8, 16, and 32. We employ FPN~\cite{FPN} following the backbone to produce 5-level features of stride 8, 16, 32, 64, and 128. The perspective head takes all 5 levels of features, while the BEV head takes the first 4 levels (with stride of 8, 16, 32, and 64).

\noindent\textbf{Perspective Head.} We adopt the single-stage anchor-free monocular 3D detector implemented by DD3D~\cite{DD3D}, which consists of three independent heads: a classification head, a 2D detection head, and a 3D detection head. 
The classification head produces the logit of each object category. 
The 2D head yields class-agnostic bounding boxes by 4 offsets from the feature location to the sides and generates the 2D center-ness. 
The 2D detection loss $\mathcal{L}_{2D}$ derives from FCOS~\cite{fcos}.
The 3D head predicts the 3D bounding boxes with the following coefficients: the quotation of allocentric orientation, the depth of the box center, the offset from the feature location to the projected box center, and the size deviation from the class-specific canonical sizes. 
Besides, the 3D head generates the confidence of the predicted 3D box relative to the 2D confidence.
It adopts the disentangles L1 loss for 3D bounding box regression and the self-supervised loss for 3D confidence in \cite{MonoDIS}, denoted as $\mathcal{L}_{3D}$ and $\mathcal{L}_{conf}$ respectively. 
The perspective loss for BEVFormer v2 is the summation of the 2D detection loss, the 3D regression loss, and the 3D confidence loss:
\begin{equation}
    \mathcal{L}_{pers} = \mathcal{L}_{2D} + \mathcal{L}_{3D} + \mathcal{L}_{conf}
\end{equation}
We refer the readers to \cite{DD3D} for more details of the perspective detection head.

\begin{figure*}[t]
    \centering
    \includegraphics[width=0.8\textwidth]{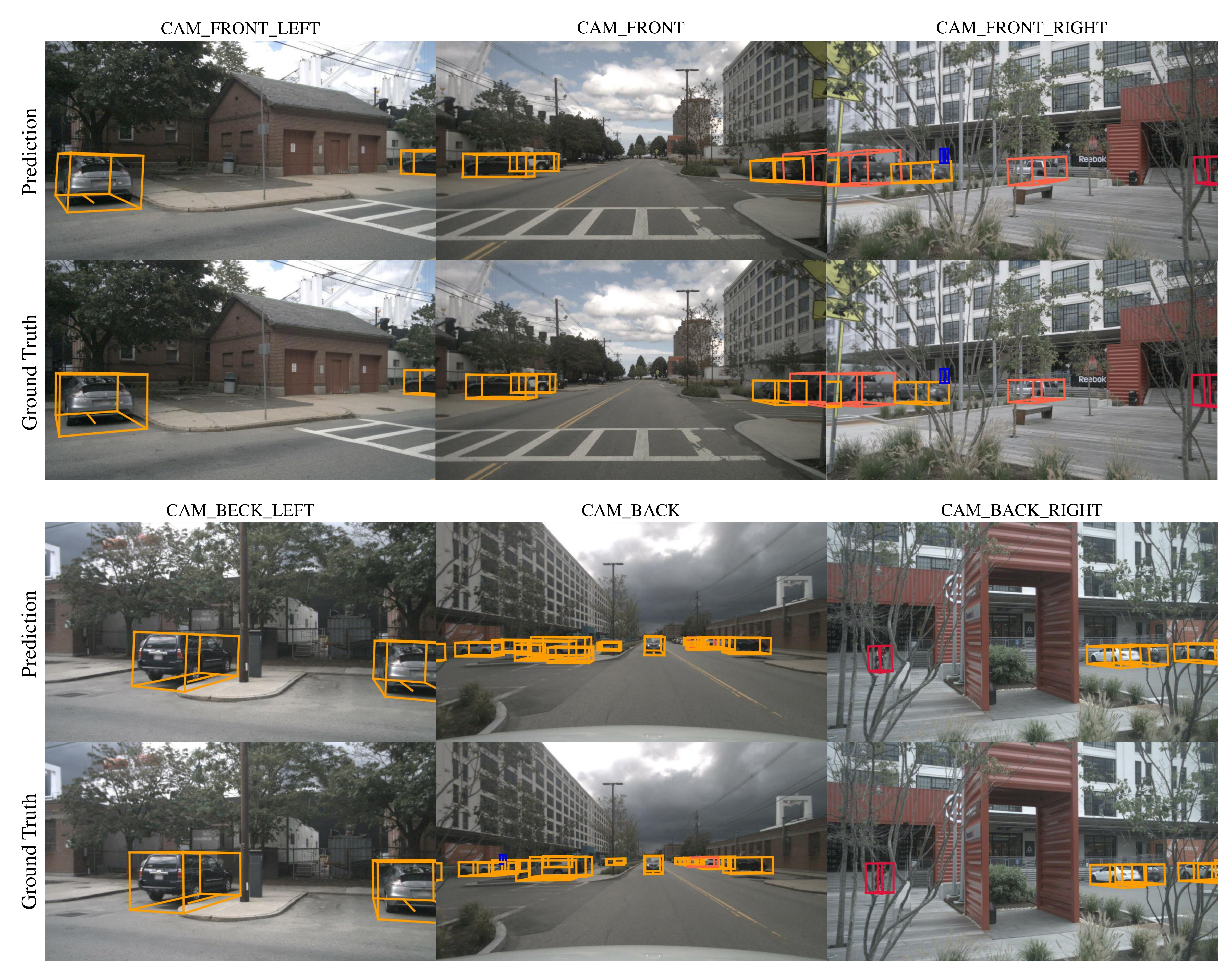}
    \caption{Visualization of BEVFormer v2 3D object detection predictions.}
    \label{fig:visualization}
\end{figure*}

\subsection{Post-Process of the First-Stage Proposals}
In this section, we describe the post-processing pipeline for proposals from the perspective detection head. 
We start with the raw predictions of all camera views provided by the perspective head. 
For the $i$-th view in all views $\mathcal{V}$, the predicted 3D bounding boxes and their scores are denoted as $\{(\mathbf{B}_{i,j},s_{i,j})\}_j$.
We filter out the candidates with the highest score (probability) through the following post-processing pipeline. 
Firstly, we perform non-maximum suppression (NMS) on the proposals of each view $i$ to obtain candidates $C_i$ without overlapping in the perspective view:
\begin{equation}
    C_i := \text{NMS}_{pers}\left(\{(\mathbf{B}_{i,j},s_{i,j})\}_j\right)
\end{equation}
The threshold of NMS is set as 2D IoU = 0.75. To ensure that objects in all camera views can be detected, we balance the numbers of proposals from different views by taking the top-$k_1$ of each view $i$ after NMS:
\begin{equation}
    C := \bigcup_{i\in\mathcal{V}} \text{top-}k_1\left(C_i\right)
\end{equation}
We set $k_1=100$ in our experiments. 
All the 3D boxes in $C$ are projected to the bird's-eye-view coordinate with corresponding camera extrinsics. 
To avoid objects that appear in multiple views causing overlapped proposals, another NMS is applied on the BEV plane with BEV IoU = 0.3: 
\begin{equation}
    C := \text{NMS}_{bev}(C)
\end{equation}
Finally, we select the top $k_2=100$ proposals: 
\begin{equation}
    C := \text{top-}k_2(C)
\end{equation}
For every 3D bounding box $\mathbf{B}$ in the final set of proposals $C$, we use its projected center on the BEV plane, $(c_x(\mathbf{B}),c_y(\mathbf{B}))$, as the reference points for Deformable DETR in the object decoder.

\section{Visualization}
We demonstrate visualization for 3D object detection results of our BEVFormer v2 detector in ~\cref{fig:visualization}. Our model predicts accurate 3D bounding boxes for the target objects, even for the hard cases in the distance or with occlusion. For instance, our model successfully detects the distant pedestrian in the front-right camera, the truck overlapped with multiple cars in the back camera, and the bicycle occluded by the tree in the back-right camera.



\end{document}